\documentclass[preprint,12pt]{elsarticle}
\usepackage{xcolor}
\usepackage{url}

\usepackage{amssymb}
\usepackage{amsmath}

\usepackage{natbib}

\usepackage{wrapfig}

\usepackage{booktabs}
\usepackage{calc}
\usepackage{tabularx}
\usepackage{array}
\usepackage{multirow}
\usepackage{caption}
\usepackage{ulem} 
\usepackage[table]{xcolor}   
\definecolor{rowgray}{gray}{0.92}

\newcommand{\R}{\ensuremath{\mathbb{R}}} 

\newcommand{\M}{\mathcal{M}}
\newcommand{\MC}{\boldsymbol{\mu}}

\newcommand{\MANAR}{MANAR }

\journal{XXXXXXXXXX}

\begin{document}

\begin{frontmatter}




\title{MANAR: Memory-augmented Attention with Navigational Abstract Conceptual Representation}


\author[l1]{Zuher Jahshan\corref{cor1}}
\ead{zuher1711@gmail.com}
\author[l1]{Ben Ben Ishay} 
\author[l1]{Leonid Yavits}
\cortext[cor1]{Corresponding author.}

\affiliation[l1]{organization={Bar Ilan University},
            city={Ramat Gan}, 
            country={Israel}}

\begin{abstract}
\textbf{MANAR} (\textbf{M}emory-augmented \textbf{A}ttention with \textbf{N}avigational \textbf{A}bstract Conceptual \textbf{R}epresentation), contextualization layer generalizes standard multi-head attention (MHA) by instantiating the principles of \textbf{Global Workspace Theory (GWT)}. While MHA enables unconstrained all-to-all communication, it lacks the functional bottleneck and global integration mechanisms hypothesized in cognitive models of consciousness. MANAR addresses this by implementing a central workspace through a trainable memory of abstract concepts and an \textbf{Abstract Conceptual Representation (ACR)}. 
The architecture follows a two-stage logic that maps directly to GWT mechanics: (i) an \textbf{integration phase}, where retrieved memory concepts converge to form a collective "mental image" (the ACR) based on input stimuli; and (ii) a \textbf{broadcasting phase}, where this global state navigates and informs the contextualization of individual local tokens. We demonstrate that efficient linear-time scaling is a fundamental architectural byproduct of instantiating GWT functional bottleneck, as routing global information through a constant-sized ACR resolves the quadratic complexity inherent in standard attention.
MANAR is a compatible re-parameterization of MHA with identical semantic roles for its projections, enabling \textbf{knowledge transfer} from pretrained transformers via weight-copy and thus overcoming the adoption barriers of structurally incompatible linear-time alternatives. MANAR enables \textbf{non-convex contextualization}, synthesizing representations that provably lie outside the convex hull of input tokens - a mathematical reflection of the creative synthesis described in GWT. Empirical evaluations confirm that MANAR matches or exceeds strong baselines across language (GLUE score of 85.1), vision (83.9\% ImageNet-1K), and speech (2.7\% WER on LibriSpeech), positioning it as an efficient and expressive alternative to quadratic attention.

\end{abstract}


\begin{highlights}
\item \textbf{Brain-Inspired GWT Architecture}: MANAR instantiates Global Workspace Theory via an Abstract Conceptual Representation (ACR) as a functional bottleneck for global information integration.
\item \textbf{Knowledge transfer and deployment}: Compatible re-parameterization of MHA enables weight-copy transfer from pretrained transformers and drop-in replacement, reducing training cost and supporting deployment in existing systems (e.g.\ document understanding, information retrieval).
\item \textbf{Linear scaling and cross-modal performance}: Linear time and memory scaling via the ACR workspace; MANAR matches or exceeds baselines on language, vision, and speech with up to 14.8$\times$ speedups and 9.3$\times$ memory reductions at scale.
\item \textbf{Non-Convex Synthesis}: MANAR produces representations outside the convex hull of input tokens, enabling creative synthesis beyond standard attention.
\end{highlights}


\end{frontmatter}



\section{Introduction}
%
Since its introduction, the transformer architecture~\cite{vaswani2017attention} has achieved remarkable success across a wide spectrum of domains, including natural language processing~\citep{vaswani2017attention, karpukhin2020dense, touvron2023llama, liu2024deepseek, warner2024smarter}, computer vision~\citep{dosovitskiy2020image, arnab2021vivit, shehzadi2023object}, speech recognition~\citep{schneider2019wav2vec, baevski2022data2vec, liu2023dinosr}, bioinformatics~\citep{brandes2022proteinbert, acera2021pacific, jahshan2024vital}, and many other domains.
At the heart of this success lies the attention mechanism, which enables every token to attend to all other tokens in a sequence, yielding highly expressive, sequence-wide contextualizations and facilitating efficient parallel training. However, the same all-to-all contextualization that powers the expressivity of multi-head attention (MHA) introduces significant scalability bottlenecks. The quadratic time and memory complexity with respect to sequence length, together with the need to store a linearly growing, unbounded context in autoregressive generation, limits both the efficiency and the reach of contemporary attention-based models. This challenge has become increasingly pressing as workloads shift toward longer contexts, higher-resolution inputs, and larger-scale models, motivating numerous architectural and algorithmic strategies to address these limitations, including hierarchical attention, recurrence, compression, and explicit memory mechanisms.

Among the most widely adopted approaches are quantization and knowledge distillation, which have succeeded in compressing the memory and compute footprint of transformer models. Quantization~\citep{ashkboos2024quarot,liu2024spinquant,liu2024kivi,xiao2023smoothquant} enables more efficient computations and reduced memory footprint by lowering precision, sometimes as low as 4-bit per element~\citep{liu2024spinquant}. Distillation~\citep{bing2025optimizing,han2024amd,mukherjee2021xtremedistiltransformers} transfers knowledge from large models to smaller ones. Despite their practical utility, these methods fall short of solving the core issue: the direct all-to-all token contextualization is preserved, so context size remains unbounded and computational complexity quadratic.


To address the unbounded-context problem, several lines of work introduce recurrence, compression, or explicit memory to extend context length without quadratic growth. Transformer-XL~\citep{dai2019transformer} introduces segment-level recurrence, allowing hidden states from previous segments to be reused as extended context while avoiding full recomputation. Building on this idea, the Compressive Transformer~\citep{rae2019compressive} augments recurrence with a compressed long-term memory, enabling retention of distant context at reduced resolution. Memory Transformer~\citep{burtsev2020memory} augments standard transformers by adding trainable memory tokens that accumulate non-local representations and help capture properties of the entire sequence beyond what element-wise contextual embeddings provide. Such memory tokens can act as dedicated slots for storing global patterns, and bottlenecks can restrict global information propagation to emphasize essential representations.

More recently, many works~\citep{xiao2024infllm, fountas2025human, sun2024shadowkv, lee2024infinigen, liu2024memlong, mohtashami2023landmark, wu2022memorizing} focus on offloading the KV cache into lower memory tiers (e.g., CPU DRAM) and sparsely selecting KV pairs to attend to. These techniques exploit the empirical observation that, during contextualization, only a small subset of tokens significantly contributes due to highly peaked attention distributions. To capitalize on this behavior, several approaches augment multi-head attention with explicit memory units that store contextual blocks (i.e., KV blocks) for later retrieval. For example, InfLLM~\citep{xiao2024infllm} contextualizes each token using both a local context window and a set of highly related KV blocks retrieved via approximate nearest-neighbor search. The retrieved blocks and the local context jointly participate in token contextualization, enabling scalable long-context inference.
However, while these strategies effectively reduce active computation time, they do not resolve the fundamental architectural burden of maintaining a linearly growing and unbounded KV cache, which remains a primary bottleneck for scaling to ultra-long sequences.

Many works seek to completely replace the standard attention mechanism with alternative architectures designed for enhanced scalability. For example, Titans~\citep{behrouz2024titans} and ATLAS~\citep{behrouz2025atlas} augment standard attention (used as short-term memory) with explicit long-term neural memory modules, which are updated using test-time training and optional persistent memory to retrieve and fuse past information alongside current-context attention for scalable long-range modeling. State space models~\citep{gu2021efficiently} formulate sequence modeling as linear dynamical systems, allowing fast, parallel computation and effective modeling of long-range context. Mamba~\citep{gu2023mamba} extends this paradigm by introducing selective sequence modeling through dynamic state selection and gating, enabling efficient and expressive representations for extremely long inputs~\citep{dao2024transformers}. Other notable advances, such as RetNet~\citep{sun2023retentive}, propose recurrent architectures with retention mechanisms that further boost performance, demonstrating strong results on long-context benchmarks. These approaches fundamentally rethink sequential model design and show particular promise in ultra-long sequence regimes.

Nonetheless, a significant practical drawback of these alternative linear-time architectures is that they fundamentally alter the parameterization of the contextualization mechanism. Because they replace the standard attention mechanism with structurally different formulations, such models cannot directly inherit or easily transfer knowledge from existing pretrained transformer attention weights. This structural incompatibility creates a substantial barrier to practical adoption, as it prevents these architectures from leveraging the vast representational knowledge already stored in large-scale pretrained models.

We introduce \MANAR\footnote{Code: \url{https://github.com/zuherJahshan/manar}}, a brain-inspired, memory-centric attention architecture that functions as a contextualization layer and can be plugged into commodity transformer encoder models. The architecture draws inspiration from cognitive processes in which perception and comprehension depend not only on sensory inputs but also on internalized concepts built from prior experience. When presented with an external input, the brain builds a mental image based on memorized concepts associated with the observed input and their relationships; this mental image then guides contextualization, the process in which meaning is assigned to each observed input occurrence.

Concretely, given an input sequence, \MANAR (i) retrieves memory concepts and constructs a full-context, constant-sized Abstract Conceptual Representation that functions as a mental image of the sequence's global themes, and (ii) contextualizes each input token using this ACR together with a local context window, avoiding direct all-to-all contextualization. \MANAR can be integrated as a drop-in replacement for standard MHA layers, making it practically deployable in existing transformer encoder stacks; weight-copy from pretrained models enables rapid adaptation by training only the additional memory-related parameters, and the design supports application areas such as information retrieval, knowledge management, and data or text mining. Empirical evaluation across language understanding, image classification, and speech recognition shows that \MANAR matches or exceeds strong baselines while delivering substantial inference speedups and peak GPU memory reductions as sequence length increases.

The MANAR architecture exhibits a deep functional connection to Global Workspace Theory (GWT)~\cite{baars1988cognitive}, which hypothesizes that the brain functions through a central workspace where information from various specialized modules is integrated and subsequently "broadcast" to the rest of the system~\cite{baars2002conscious}. While standard multi-head attention (MHA) allows for all-to-all communication, it lacks the functional bottleneck that GWT suggests is necessary for coherent global integration~\cite{dehaene1998neuronal}. MANAR implements this workspace via the Abstract Conceptual Representation (ACR), which serves as a unified mental image that guides the contextualization of local inputs.



\section{Preliminaries and background}

We begin by formalizing the notions of \emph{concept} and \emph{contextualization}, which we use throughout this work to reason about attention mechanisms and to motivate the \MANAR architecture. These notions are inspired by cognitive models of human perception as well as by the operational structure of modern attention-based neural networks.

\paragraph{Concept}
We model the fundamental unit of information processed by attention layers as a \emph{concept}. A concept encapsulates not only semantic content but also the mechanisms by which this content interacts with other information. Concretely, a concept is represented as a triplet $(q,k,v)$, where the \emph{query} $q$ determines how the concept seeks relevant context, the \emph{key} $k$ determines how it contributes to the contextualization of other concepts, and the \emph{value} $v$ represents the semantic content carried by the concept.
In the GWT framework, individual tokens and memorized concepts act as specialized resources competing for representation within the global workspace~\cite{dehaene2001towards}. The ACR acts as the functional bottleneck described in cognitive models, ensuring that global information is integrated into a compact, constant-sized representation before influencing the wider network.

\paragraph{Contextualization}
Contextualization is the process by which the meaning of a concept is refined through interaction with other concepts. Given a concept $x=(q,k,v)$ and a set of concepts $C=\{(c_1^q,c_1^k,c_1^v),\dots,(c_n^q,c_n^k,c_n^v)\}$, contextualization produces an updated representation by aggregating the values of concepts in $C$ according to their relevance to $x$. Relevance is measured via a similarity function between the query of $x$ and the keys of the contextualizing concepts.

Formally, the contextualized representation $y$ of $x$ with respect to $C$ is given by
\[
y = \sum_{(c^q,c^k,c^v)\in\tilde{C}} S(q,c^k)\,c^v,
\quad \text{where } \tilde{C}=C\cup\{x\},
\]
with $S(\cdot,\cdot)\ge0$ and $\sum_{(c^q,c^k,c^v)\in\tilde{C}} S(q,c^k)=1$.

This formulation implies that contextualization produces a convex combination of value vectors. Consequently, the contextualized representation is constrained to lie within the convex hull spanned by the values of the contextualizing concepts.

\begin{wrapfigure}{r}{0.42\columnwidth}
\centering
\includegraphics[
    width=\linewidth,
    clip
]{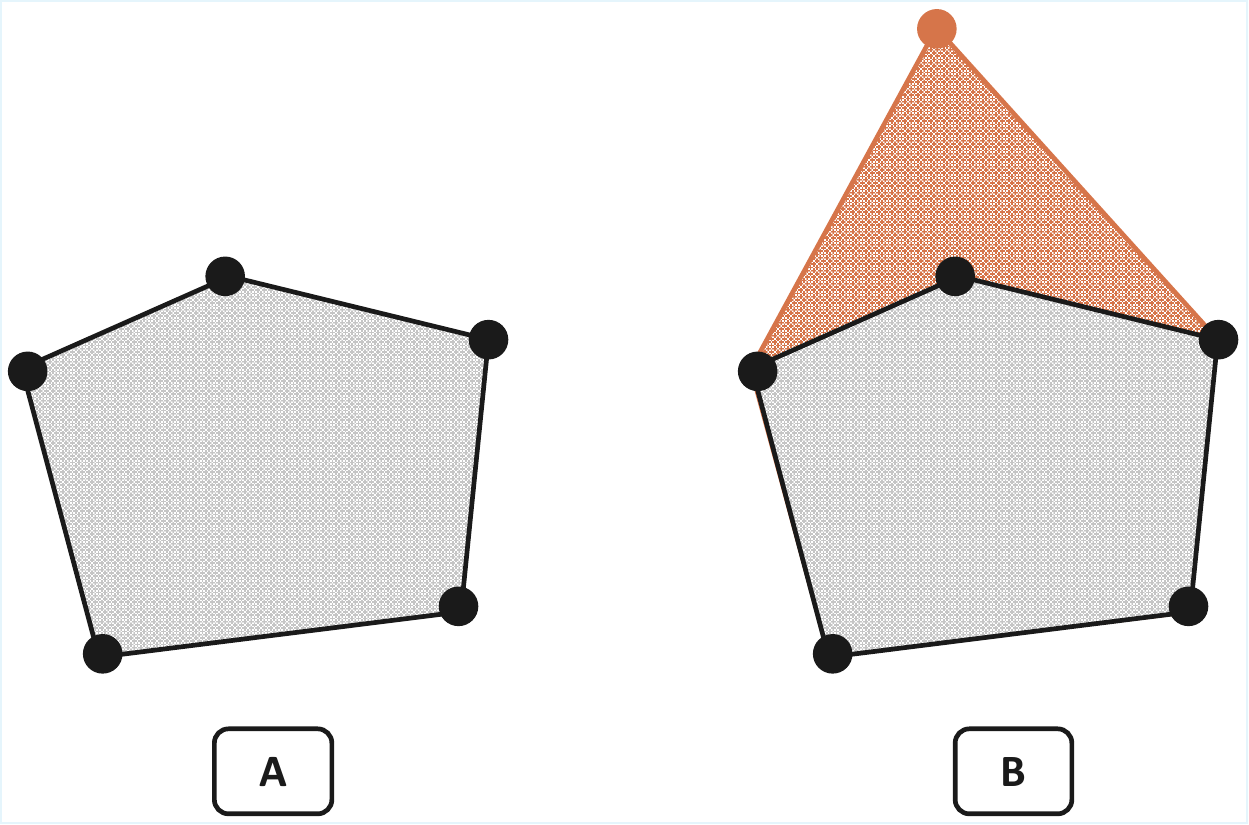}
\caption{Geometric interpretation of contextualization with and without retrieved memory concepts.}
\label{fig:OOTB}
\end{wrapfigure}

This geometric interpretation highlights a fundamental limitation of standard contextualization mechanisms: when contextualization is restricted to the input sequence alone, the expressivity of the output is bounded by the convex hull of the input token values, as illustrated in Fig.~\ref{fig:OOTB}(A).

However, when contextualization incorporates \emph{retrieved memorized concepts} that are not present in the input sequence, this constraint can be relaxed. As shown in Fig.~\ref{fig:OOTB}(B), the inclusion of external concept values expands the reachable region in representation space. We refer to this capability as \emph{non-convex contextualization}, since it enables the synthesis of representations that cannot be obtained by reweighting the input tokens alone.

\paragraph{Multi-head attention as contextualization}
The multi-head attention (MHA) mechanism can be interpreted as a particular instantiation of the contextualization process described above, where each input token is conceptualized by a learned $(q,k,v)$ triplet and contextualized exclusively by other input tokens in the sequence.

In this work, we aim to develop a different contextualization process. When an external sequence of inputs is presented, internal memorized concepts that are strongly associated with the input are retrieved (an idea inspired by the human brain). Links and associations are then formed between input tokens and retrieved memorized concepts. These links take the shape of a constructed global conceptual representation of the perceived input. This constructed representation then guides and navigates the contextualization of the presented inputs.
We can examine human reading comprehension as an example of this behavior. When presented with a text, the words stimulate the brain to retrieve memorized concepts associated with it. Throughout the reading process, the brain constructs a mental image, an abstract conceptual representation, capturing the perceived meaning of what is being read and its connections to memory. Each word is then contextualized and understood in light of this constructed abstract representation~\citep{kewenig2024abstract, keller2024neural}.

\section{\MANAR}
\label{sec:manar}

\MANAR contextualizes input tokens by neighboring tokens as well as by retrieved memorized concepts not present in the input sequence. To make this possible, \MANAR integrates a memory unit that retains memorized concepts. When an input sequence of tokens $X$ is presented, $m$ search patterns are generated and applied to retrieve $m$ memorized concepts from the memory unit in a fast and scalable manner. After retrieval, these internal memory concepts are linked and associated to the presented input forming the Abstract Conceptual Representation ($ACR$). 
Then, input tokens are contextualized by the $ACR$ as well as the local context window of the token to formulate the output of the layer.

\subsection{Notations}
We start by discussing some notations that are used consistently throughout the paper. Let $X\in\R^{n\times D}$ represent the input sequence of tokens, where $n$ is the sequence length and $D$ is the dimension of each observed input token. Let $\M_i$ be the $i$-th retrieved memory concept which is consistently represented as a qkv-tuple $\M_i=(c_i^q,c_i^k,c_i^v)$. Use $m$ to refer to the number of retrieved memory concepts, and $M$ to refer to the total memory size (i.e., the total number of memory cells in the memory unit). The $ACR$ size is $m\times d$ (where $d$ is the per-head dimension), a row for each retrieved memory concept. We refer to the $i$-th row of the $ACR$ by $ACR_i$ or $r_i$ interchangeably. We refer to $d$ as the intra-layer dimension (i.e., per-head dimension), and we assume $D=hd$ where $h$ is the number of heads. Lastly, the $i$-th output (i.e, the contextualized $i$-th token) is referred to as $y_i$. In this work we follow a row-major representation.

\subsection{ACR construction and Token Contextualization}
\label{sec:ootb}
Throughout Sec.~\ref{sec:ootb} we assume the existence of $m$ retrieved memory concepts $\big\{\M_i=(c_i^q,c_i^k,c_i^v)\big\}_{i=1}^{m}$ decoupling the process of memory retrieval from the rest of the \MANAR architecture. Memory retrieval is discussed separately in Sec.~\ref{sec:mem-ret}. Calculations are made considering a single-head architecture.
Multi-head architecture generalization is made in Section~\ref{sec:mh_manar}.

\MANAR defines four learnable projection matrices $W_q,W_{k}^{\M},W_{k},W_v\in \R^{D\times d}$ corresponding to the token's query, ACR key, contextualization key, and value, respectively. $W_k^{r}\in\R^{d\times d}$ represents the projection responsible for converting $ACR$'s into "token contextualization" key-space. Moreover, we define the Region of Interest of an index $i$ with a neighborhood $l$ as:
\begin{align*}
&ROI^l(i)=\{j: \max(0,i-l+1) < j\le \min(n,i+l)\}
\end{align*}

where $ROI^l(i)[j]$ represents the $j$-th smallest element in the set. The $ROI$ is used to represent the local context window in the process of token contextualization. We refer to the local context window length as the maximal size of $ROI^l$, which is $2l$.
Equipped with these learnable parameters and the $ROI$, the logic of the \MANAR layer operating in two stages (i.e., the Integration and Broadcasting stages), the ACR construction and the token contextualization, is defined as follows:
\begin{itemize}
    \item{Conceptualization:}
    \begin{align}
    \label{eq:concept}
    k_i^{\M} = x_iW^{\M}_{k};&&
    q_i = x_iW_q;&&
    k_i = x_iW_{k};&&
    v_i = x_iW_v
    \end{align}
    \item{Integration stage (ACR Construction): }\begin{align}
        & r_i = S_{i,0}c_i^v + \sum_{j=1}^{n}{S_{i,j}v_j} \label{eq:acr} \\
        \text{where   } & (S_{i,0},S_{i,1}, S_{i,2},\dots, S_{i,n}) =\\& softmax\Bigg(\frac{c_i^q \cdot (c_i^k)^T}{\sqrt{d}}, \frac{c_i^q\cdot (k_1^{\M})^T}{\sqrt{d}}, \dots, \frac{c_i^q\cdot (k_n^{\M})^T}{\sqrt{d}}\Bigg) \nonumber
    \end{align}
\end{itemize}
The process of ACR construction can be seen as contextualizing each retrieved memory concept by the conceptualized input tokens. Concretely, the $i$-th ACR vector, $r_i$, represents the meaning of the memorized concept shifted according to how strongly that memory concept associates with each observed token. The strength of an association is measured by the inner product $c_i^q\cdot(k_j^{\M})^T$. Intuitively, this mirrors human cognition: our mental image of a situation blends internal memorized concepts with incoming evidence, weighted by the perceived relevance of each piece of evidence to those concepts.

\begin{itemize}
    \item{Broadcasting stage (Token Contextualization): }\begin{align}
            & y_i = \underbrace{\sum_{j=1}^{m}{\hat{S}_{i,j}r_j}}_{\text{global attention}}+\underbrace{\sum_{j=1}^{L}{\tilde{S}_{i,j}v_{ROI^l(i)[j]}}}_{\text{local attention}} \label{eq:moox-resp}\\
        \text{where } & (\hat{S}_1,\dots,\hat{S}_m,\tilde{S}_{1},\dots,\tilde{S}_{L})=\\
        &softmax\Bigg(\frac{q_i(r_1W_k^r)^T}{\sqrt{d}},\dots,\frac{q_i(r_mW_k^r)^T}{\sqrt{d}},\frac{q_ik_{ROI^l(i)[1]}^T}{\sqrt{d}},\dots,\frac{q_ik_{ROI^l(i)[L]}^T}{\sqrt{d}}\Bigg) \nonumber\\
        \text{and } L&=|ROI^l(i)|\nonumber
    \end{align}
\end{itemize}
Navigated by the $ACR$, \MANAR contextualizes each input token in light of this global representation as well as all the input information that could be perceived at once (i.e., the local context window), represented by the $ROI$ of the token. Hence, the meaning each contextualized token holds is influenced both by association with the $ACR$ and with neighboring tokens.

Since the ACR is constructed around memorized concepts and the meaning they hold, \MANAR-contextualized tokens can obtain meanings not bounded by the meaning space spanned by the convex hull of inputs $\{xW_v:x=X_i,1\le i\le n\}$.
We refer to this behavior as \textit{non-convex contextualization}. This effect is expressed by the following derivation of $r_i$:

\begin{align}
    r_i&=S_{i,0}c_i^v + (1-S_{i,0})\sum_{j=1}^{n}{\frac{S_{i,j}}{1-S_{i,0}}v_j}\\
    &=S_{i,0}c_i^v+(1-S_{i,0})\sum_{j=1}^{n}{\frac{\frac{e^{c_i^q(k_j^m)^T}}{e^{c_i^q(c_i^k)^T}+\sum_{l=1}^{n}{e^{c_i^q(k_l^m)^T}}}}{\frac{\sum_{l=1}^{n}{e^{c_i^q(k_l^m)^T}}}{e^{c_i^q(c_i^k)^T}+\sum_{l=1}^{n}{e^{c_i^q(k_l^m)^T}}}}v_j}\\
    &=S_{i,0}\underbrace{c_i^v}_{\text{B}} + (1-S_{i,0})\underbrace{\sum_{j=1}^{n}{\frac{e^{c_i^q(k_j^m)^T}}{\sum_{l=1}^{n}{e^{c_i^q(k_l^m)^T}}}v_j}}_{\text{A}} \label{eq:tc}
\end{align}
Eq.~\ref{eq:tc} demonstrates that each ACR row $r_i$ is a weighted sum of two terms: (A) an expression having the same form as the output of MHA; (B) a correction term that can shift $r_i$ outside the convex hull of the input token values, as illustrated in Fig.~\ref{fig:OOTB}(B).

The two-stage logic of MANAR maps directly to the mechanics of a global workspace. Stage 1 (ACR Construction) represents the integration phase, where relevant memorized concepts are retrieved and shifted based on their association with the input sequence to form a collective "mental image". Stage 2 (Token Contextualization) corresponds to the broadcasting phase, where the global state maintained in the ACR navigates and informs the understanding of each individual token. This ensures that local perception is interpreted in light of the global context, mirroring the cognitive process where the workspace broadcasts content to inform specialized processing~\cite{baars2002conscious}.

\subsection{The Memory Unit}
\label{sec:mem-ret}
In this section, we discuss the memory unit and the memory retrieval process of $m$ concepts, $\M_i=(c_i^q,c_i^k,c_i^v)$, completing the full picture of \MANAR. 

The memory unit contains $M$ memory cells. Each memory cell retains a concept, $\MC_i=(\mu_i^q,\mu_i^k,\mu_i^v)$, where $0< i \le M$. The memory retrieval process
involves the creation of $m$ different search patterns as a function of input tokens. For each search pattern, top-\textit{k} memory cells are chosen on the basis of their similarity to the search pattern. The memory concept is calculated as a weighted sum of the contents of these matching top-\textit{k} memory cells. The logic of producing the search pattern is first formalized, then we detail how each search pattern drives retrieval. 

To perform $m$ memory lookups, the model first constructs $m$ search patterns from the input sequence $X\in \R^{n\times D}$. Concretely, the model introduces $m$ learnable "mixer" vectors $mixer_i\in\R^d$ that aggregate information from tokens via a cross-attention operation where the queries are the mixer vectors and the keys/values come from the tokens. Let $W_k^{SP},W^{SP}_v\in\R^{d\times d}$ be learnable projections; the $i$-th search pattern is then:
\begin{align}
\label{eq:spb}
    \sigma_i=softmax\Bigg(\frac{mixer_i\cdot (XW_k^{SP})^T}{\sqrt{d}}\Bigg)\cdot XW^{SP}_v
\end{align}

Given a search pattern $\sigma_i$, the memory unit keys, a table of keys, one per each memory cell, $\xi\in\R^{M\times d}$, and the memory cells $\MC\in\R^{M\times3d}$, retrieval computes a soft combination of cells weighted by their similarity to $\sigma_i$. The retrieval step is: 
\begin{align}
    \label{eq:mcr}
    &I=SelectTopkIndices(\sigma_i\cdot\xi^T);&&
    &s=softmax(\sigma_i\cdot(\xi_I)^T);&&&
    &\M_i=s\cdot\MC_I
\end{align}
where $I$ is a set of indices, $s\in\R^k$, $\xi_I\in\R^{k\times d},\MC_I\in\R^{k\times (3d)}$, and the output $\M_i\in\R^{3d}$.

Increasing the memory size, $M$, makes naive nearest-neighbor scoring over all keys $\mathcal{O}(M)$ per search pattern prohibitive; fast approximate similarity search techniques could be used here~\citep{johnson2019billion}, but incorporating them is challenging when keys are continually trained and re-indexed. 
\MANAR uses trainable product keys~\citep{lample2019large}, which factor the key space into two tables $\xi^{(1)},\xi^{(2)}\in\R^{\sqrt{M}\times \frac{d}{2}}$ whose (implicit) Cartesian product spans M composite keys without materializing them. For lookup, split the search pattern as $\sigma_i=\big[\sigma_i^{(1)};\sigma_i^{(2)}\big]$ with $\sigma_i^{(1)},\sigma_i^{(2)}\in\R^{\frac{d}{2}}$, retrieve top-k indices and scores $(I_1,s_1)$ from $\xi^{(1)}$ and $(I_2,s_2)$ from $\xi^{(2)}$, then combine candidates by maximizing summed scores over pairs:
$$argmax_{j_1\in I_1,j_2\in I_2}s_1[j_1]+s_2[j_2]
$$
which yields an efficient approximation to top-\textit{k} over the full $M$ composite keys while searching only the $\sqrt{M}$-sized half-key tables.

\begin{figure}
    \centering\includegraphics[width=10cm]{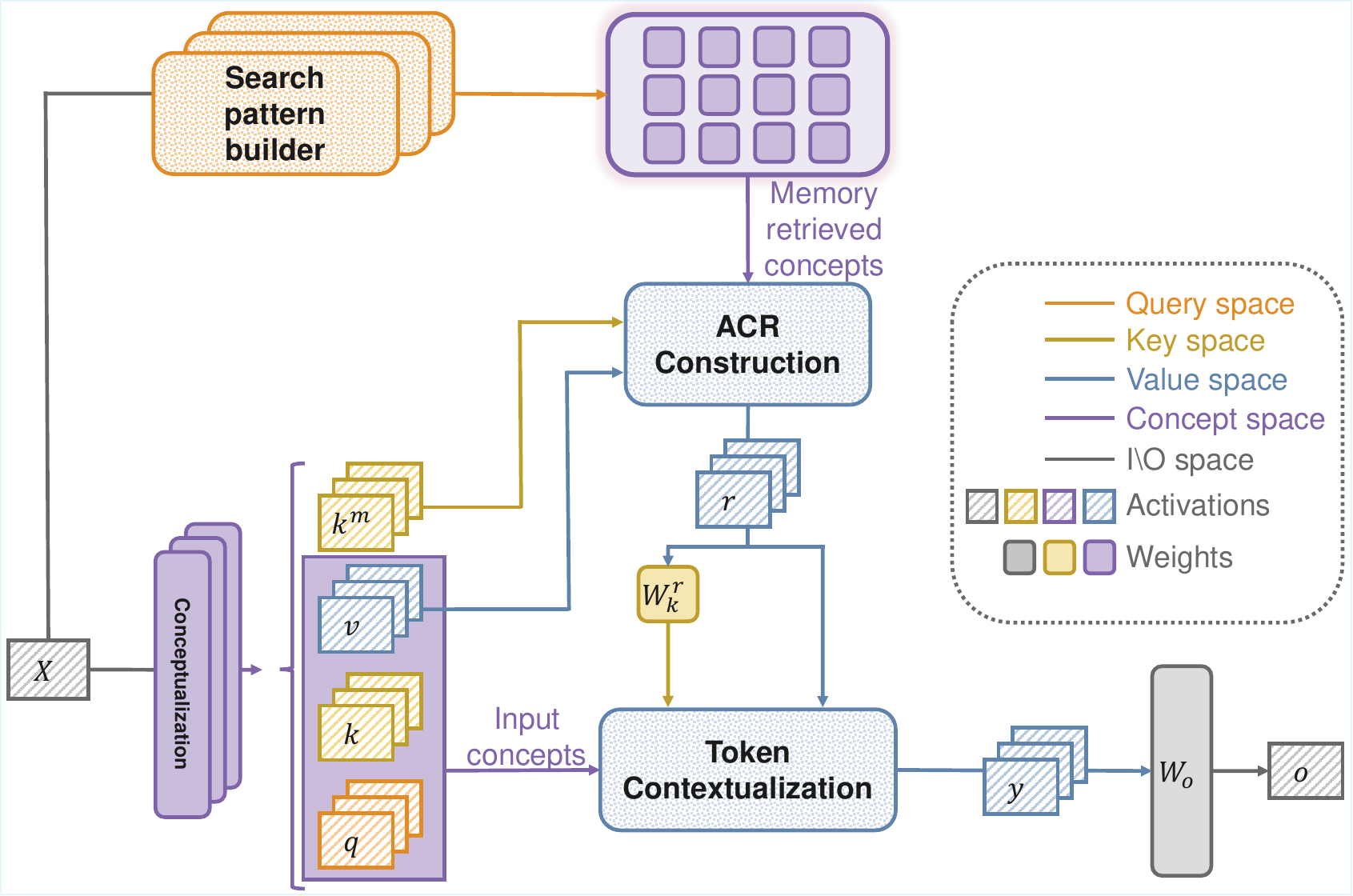}
    \caption{High level architecture of \MANAR}
    \label{fig:arch}
\end{figure}

\subsection{Multi-head architecture}
\label{sec:mh_manar}
To form a $h$-head \MANAR layer we replicate the entire single-head conceptualization pipeline, including the search-pattern builder, one per head. Each head therefore learns its own token projections and search patterns, yet all heads reuse the same token-contextualization key projection $W_k^r$ and access one shared external memory. After each head completes retrieval, ACR construction and token contextualization, their outputs are concatenated and mapped back to $D$ dimensions through a single output projection, preserving the standard transformer interface.

\section{Evaluation}
\label{sec:eval}
To evaluate \MANAR accuracy, performance, and memory usage, we apply it as a drop-in replacement to MHA in several transformer-based models, corresponding to language, image, and speech modalities.
The chosen benchmarks are representative of application settings: e.g.\ search and document analysis, visual recognition, and speech interfaces.

Across all modalities, our primary goal is to isolate the effect of the \emph{contextualization mechanism}. Unless explicitly stated otherwise, our comparisons follow the same contract: we keep the backbone architecture, training pipeline, and evaluation procedure fixed, and change only the contextualization layer (MHA $\rightarrow$ \MANAR).
All models we compare are parameter-budget matched (all base-sized models are kept below 150M parameters).
We report wall-clock inference comparisons in Sec.~\ref{sec:perfcomp} (Performance Comparison).

Because training recipes are highly non-unique and domain-specific, we do not claim that every baseline is fully optimized for best possible accuracy under all settings. Instead, we focus on controlled, apples-to-apples runs where \MANAR and the baseline share the same data, preprocessing, optimizer, schedule, and evaluation protocol, so that observed differences can be attributed to the contextualization mechanism.

When comparing vanilla transformer encoder architecture to \MANAR-enabled one, we leave all other layers unmodified. In this section, we refer to any \MANAR-enabled transformer encoder architecture, having a memory of size $M$ and $ACR$ of size $m$ that aggregates top-$8$ memory cells to assemble a memory concept, with a context window length of $C$ as \texttt{MANAR-M.m.C}.

\subsection{Language Modeling}
\label{sec:lm}
We evaluate \MANAR on natural language understanding via masked language model (MLM) pre-training followed by fine-tuning on downstream GLUE tasks.
The purpose of this experiment is to isolate the impact of replacing multi-head attention (MHA) with \MANAR in a BERT-base–style encoder, under a controlled RoBERTa-like~\citep{liu2019roberta} training procedure and using absolute positional embeddings. We exclude Rotary Positional Embeddings (RoPE) in this section to avoid conflating architectural effects with positional encoding choices.

Following the controlled protocol of Sec.~\ref{sec:eval}, we replace only MHA with \MANAR while keeping all other encoder components and the training pipeline identical.

For controlled comparisons, we train two families of models from scratch on the English subset of C4 using the MosaicML~\citep{portes2023mosaicbert} framework, following the RoBERTa recipe: Next Sentence Prediction is removed, dynamic MLM masking is applied with a 15\% masking probability, the sequence length is 512 tokens, training runs for 30K steps (i.e., only a portion of the C4 English subset accounting for 135\,GB of data) with a global batch size of 4K, and all models use the \texttt{bert-base-uncased} tokenizer (30{,}522 tokens). Optimization uses AdamW with peak learning rate $5\times10^{-4}$ and the linear warmup with linear decay learning schedule~\citep{devlin2019bert,baevski2022data2vec,liu2019roberta,portes2023mosaicbert}. All trained models use the same data subset and identical optimization and schedule settings. Within this controlled regime, we train (i) a RoBERTa baseline, and (ii) two \MANAR variants that act as strict drop-in replacements for MHA while keeping all other encoder components unchanged.

In addition to these controlled runs, we also report the published GLUE results of BERT~\cite{devlin2019bert} and data2vec~\citep{baevski2022data2vec} in Table~\ref{tab:lm}. We include BERT as a historically standard BERT-base reference point (trained under earlier, smaller-data regimes) and data2vec as a strong modern self-supervised baseline that achieves competitive GLUE performance. Importantly, these two rows are taken \emph{as reported} and are not retrained; the controlled, apples-to-apples comparison in our setting is therefore between RoBERTa run and \MANAR runs.

The first variant, \texttt{MANAR-484.64.128}, uses a memory of 484 concepts, retrieves $m=64$ concepts to form the ACR, and uses a local context window length of $C=128$. The larger-memory configuration, \texttt{MANAR-16K.128.128$^\dagger$} uses a hybrid architecture: every third encoder layer uses a memory size of 16K concepts, retrieving $m=128$ concepts to form the ACR, while the remaining layers use an identical configuration as of the smaller configuration (i.e., $M=484$, $m=64$, and $C=128$).
This hybrid design increases memory capacity while keeping overall compute and parameters in the BERT-base scale (i.e. $<150M$ parameters).

Table~\ref{tab:lm} reports GLUE results after fine-tuning. Under the controlled training regime (RoBERTa vs.\ \MANAR variants), \MANAR improves performance broadly across tasks and achieves the strongest average score among the models trained in this setting. The gains become larger as memory capacity increases: moving from \texttt{MANAR-484.64.128} to \texttt{MANAR-16K.128.128$^\dagger$} yields consistent improvements across nearly all GLUE tasks, 
demonstrating that accuracy improves monotonically with memory size, as a larger memory enables richer retrieval of abstract concepts for contextualization.

Finally, while newer encoders~\citep{warner2024smarter,weller2025seq} can reach higher absolute GLUE numbers, they typically differ from our setup by adopting RoPE, longer-context training regimes, and much larger pre-training data. Because our goal here is to attribute changes in downstream quality specifically to the contextualization mechanism, we leave RoPE integration and longer-context pre-training for future work.

\begin{table*}[t]
\centering
\caption{GLUE development set results (\%). Higher is better.}
\label{tab:lm}

\setlength{\tabcolsep}{6pt}
\renewcommand{\arraystretch}{1.15}
\small

\resizebox{\textwidth}{!}{%
\begin{tabular}{lccccccccc}
\toprule
\textbf{Model} &
\textbf{MNLI} &
\textbf{QNLI} &
\textbf{RTE} &
\textbf{MRPC} &
\textbf{QQP} &
\textbf{\mbox{STS-B}} &
\textbf{CoLA} &
\textbf{SST} &
\textbf{Avg.} \\
\midrule
BERT~\citep{devlin2019bert}
& \uline{84.0}/\uline{84.4} & 89.0 & 61.0 & 86.3 & 89.1 & 89.5 & 57.3 & 93.0 & 80.7 \\
data2vec~\citep{baevski2022data2vec}
& 83.2/83.0 & \uline{90.9} & 67.0 & 90.2 & 89.1 & 87.2 & \uline{62.2} & 91.8 & 82.7 \\
RoBERTa~\citep{liu2019roberta}
& \textbf{84.4}/84.1 & 90.4 & \textbf{75.3} & 89.8 & \uline{89.6} & 89.3 & 57.4 & 92.3 & 83.4 \\
\rowcolor{rowgray}
\texttt{MANAR-484.64.128}
& 83.7/83.4 & 90.3 & 71.9 & \uline{90.6} & 89.1 & \uline{90.3} & 59.7 & \textbf{94.4} & \uline{83.8} \\
\rowcolor{rowgray}
\texttt{MANAR-16K.128.128}$^\dagger$
& \uline{84.0}/\textbf{84.6} & \textbf{91.2} & \uline{73.6} & \textbf{91.7} & \textbf{90.2} & \textbf{90.5} & \textbf{63.1} & \uline{94.1} & \textbf{85.1} \\
\bottomrule
\end{tabular}%
}

\vspace{2pt}
\caption*{\footnotesize
MNLI is reported as matched/mismatched (m/mm). Best results are in \textbf{bold}; second-best are \uline{underlined}. $^{\dagger}$ Models where every third layer uses a memory size of 16K; all other layers use a memory size of 484, and an ACR size of 64.
}
\end{table*}

\subsection{Image Classification}
\label{sec:image_training}
Furthermore, we benchmark \MANAR on the ImageNet-1K dataset~\citep{deng2009imagenet}, which contains 1.28M training images and 50K validation images from 1000 categories. We use DeiT-B and DeiT-S~\citep{touvron2021training} as our baselines, hence, we refer to \texttt{MANAR-M.m.C-B(-S)} as a DeiT-B(-S) transformer encoder where all MHA layers were replaced by \MANAR layers. We trained a \texttt{MANAR-256.32.96-B(-S)} model with 12 encoder layers, each containing 12(6) heads. The dimensionality of each head was set to $d=64$.

As in all experiments, only the attention blocks are replaced (MHA $\rightarrow$ \MANAR); the backbone, augmentation, optimizer, and schedule remain unchanged.
Timing and memory comparisons are reported in Sec.~\ref{sec:perfcomp}.
The model is trained on the training set and the top-1 accuracy on the validation set is reported. For fair comparisons, we trained the model from scratch with training settings used in DeiT. Specifically, we apply random cropping, random horizontal flipping, label smoothing regularization, mixup, and random erasing as data augmentations. The training took place on images of size $224^2$. We employ AdamW~\citep{loshchilov2017fixing} with $\beta_1{=}0.9$, a total batch size of 1024, and a weight decay of $5\cdot 10^{-2}$ to optimize the model. We train the \MANAR-based DeiT architecture for 450(300) epochs using the cosine scheduling with a learning rate initiated as $4\cdot 10^{-4}$ and Exponential Moving Average (EMA). During testing we apply a center crop on the validation set to crop out $224^2$ images. Experiments are performed on a single H100 GPU. Top-1 accuracy on validation set results are reported in Tab.~\ref{tab:imagenet}. We compare against DeiT-B(-S)~\citep{touvron2021training}, Vision Mamba Vim-B(-S)~\citep{zhu2024vision}, a linear-complexity architecture, and the vanilla Vision Transformer~\citep{dosovitskiy2020image}. As Table~\ref{tab:imagenet} shows, \MANAR consistently outperforms models of comparable size, confirming its ability to achieve competitive accuracy while operating at linear complexity. 

\begin{table*}[t]
\centering
\caption{Comparison of different backbone architectures on ImageNet-1K.}
\label{tab:imagenet}

\setlength{\tabcolsep}{6pt}
\renewcommand{\arraystretch}{1.05}
\footnotesize

\begin{tabular}{lcc}
\toprule
\textbf{Model} & \textbf{Image Size} & \textbf{Top-1 Acc. (\%)} \\
\midrule

\multicolumn{3}{c}{\textbf{Small Models}} \\
\midrule
DeiT-S~\citep{touvron2021training} & $224^2$ & 79.8 \\
Vim-S~\citep{zhu2024vision}       & $224^2$ & 80.3 \\
\rowcolor{rowgray}
\texttt{MANAR-256.32.96-S}                & $224^2$ & \uline{80.7} \\
\rowcolor{rowgray}
\texttt{MANAR-4K.32.96-S}$^{\dagger}$     & $224^2$ & \textbf{81.6} \\

\midrule

\multicolumn{3}{c}{\textbf{Base Models}} \\
\midrule
ViT-B/16~\citep{dosovitskiy2020image} & $384^2$ & 77.9 \\
ViT-L/16                             & $384^2$ & 76.5 \\
DeiT-B                               & $224^2$ & 81.8 \\
Vim-B~\citep{zhu2024vision}           & $224^2$ & 81.9 \\
\rowcolor{rowgray}
\texttt{MANAR-256.32.96-B}                    & $224^2$ & \uline{82.3} \\
\rowcolor{rowgray}
\texttt{MANAR-4K.128.96-B}$^{\dagger}$        & $224^2$ & \textbf{83.9} \\

\bottomrule
\end{tabular}

\vspace{2pt}
\footnotesize
\raggedright
$^{\dagger}$ Models where every third layer uses a memory size of 4K; all other layers use a memory size of 256.
\end{table*}

\subsection{Knowledge Transfer}
\label{sec:kt}
A defining property of \MANAR is that it serves as a compatible re-parameterization of MHA, exposing \texttt{q}, \texttt{k}, \texttt{v}, and \texttt{out\_proj} projection matrices with the same semantic roles as in standard attention. As discussed in Sec.~1, alternative linear-time architectures fundamentally alter the parameterization of contextualization and therefore cannot directly inherit pretrained weights. To our knowledge, \MANAR is the only linear-complexity contextualization mechanism that supports direct weight-copy initialization from standard MHA, making it uniquely positioned for practical adoption in existing transformer ecosystems.

Concretely, knowledge transfer from a pretrained transformer to \MANAR is performed by copying the \texttt{q}, \texttt{k}, \texttt{v}, and \texttt{out\_proj} matrices from each MHA layer into the corresponding \MANAR layer. During the initial transfer phase, these copied weights are frozen, and only the newly introduced memory-related parameters are trained. This procedure preserves the inductive biases and representational structure learned by the original model, while allowing \MANAR to augment contextualization through its external memory.

We evaluate this transfer mechanism on three representative domains: language understanding, image classification, and automatic speech recognition.

For language understanding, we start from a pretrained RoBERTa model that achieves 83.4\% average GLUE after fine-tuning. All MHA layers are replaced with \MANAR layers (\texttt{MANAR-484.64.128}, i.e., $M{=}484$, $m{=}64$, $C{=}128$) initialized by copying the RoBERTa attention weights. The model is then trained for 5K MLM pre-training steps: the copied weights are frozen for the first 3K steps, after which all weights are rendered trainable for the remaining 2K steps. After fine-tuning, this transferred model achieves 83.5\% average GLUE, slightly exceeding the source model while requiring only a fraction of the 30K-step from-scratch training budget used in Sec.~\ref{sec:lm}.

For image classification, we start from a pretrained DeiT model achieving 83.4\% top-1 accuracy on ImageNet-1K. All MHA layers are replaced with \MANAR layers initialized by copying the DeiT attention weights. In the baseline transfer configuration, each \MANAR layer uses a memory size of $M{=}256$, an ACR size of $m{=}32$, and a context window length of $C{=}96$, while all copied weights remain frozen. Training only the newly introduced parameters for 20 epochs yields 83.1\% top-1 accuracy, despite updating only a small fraction of the model parameters. Compared to training \MANAR from scratch for 450 epochs, this represents a substantial reduction in training cost relative to the 450-epoch from-scratch regime.

To examine whether transferred models can further benefit from increased memory capacity, we extend this experiment by enlarging the memory in every third layer to $M{=}4\text{K}$ with $m{=}128$ and $C{=}96$, while keeping all other layers identical to the small-memory configuration. Training proceeds for a total of 50 epochs: 20 epochs with copied weights frozen, followed by 30 epochs with all weights rendered trainable. This configuration achieves 83.7\% top-1 accuracy, surpassing the original DeiT baseline. These results demonstrate that knowledge transfer in \MANAR is not only effective, but also compatible with subsequent memory enlargement, allowing transferred models to exceed the performance of their source architectures.

A parallel study is conducted for automatic speech recognition using data2vec-base as the source model.

Following the same controlled protocol, we report WER under a consistent decoding setup (including the same language model) across all compared models.

As in the vision setting, attention weights are copied into \MANAR to initialize the model. A local context window of $C{=}128$ is employed, corresponding to approximately 2.5 seconds of audio. The model is trained on 100 hours of LibriSpeech using the CTC loss, with all weights rendered trainable during training. In the baseline configuration with memory size $M{=}256$ and ACR size $m{=}64$, \MANAR matches the performance of state-of-the-art self-supervised speech models. Increasing the memory capacity to $M{=}4\text{K}$ with $m{=}128$ in every third layer yields consistent improvements across all evaluation splits, achieving 2.7\% / 6.4\% WER on the LibriSpeech test-clean / test-other sets (Table~\ref{tab:asr}), matching or exceeding the strongest published baselines.

Taken together, these results establish that \MANAR supports efficient, high-fidelity knowledge transfer from pretrained transformers across modalities. Unlike architectures that replace attention with fundamentally different mechanisms, \MANAR preserves attention’s parameterization while extending it with an expandable memory. This compatibility enables rapid adaptation, drastic reductions in training cost, and continued performance gains as memory capacity grows, making \MANAR a practical and deployable alternative to standard multi-head attention.

\begin{table*}[t]
\centering
\caption{Word Error Rate (WER; \%) on LibriSpeech standard dev/test sets. All models use the same 12-layer Transformer encoder. Decoding uses the official 4-gram language model~\citep{heafield2011kenlm}. Lower is better.}
\label{tab:asr}

\setlength{\tabcolsep}{7pt}
\renewcommand{\arraystretch}{1.15}
\footnotesize

\begin{tabular}{lcccc}
\toprule
\textbf{Model} & \textbf{dev-clean} & \textbf{dev-other} & \textbf{test-clean} & \textbf{test-other} \\
\midrule
wav2vec2.0~\citep{baevski2020wav2vec}   & 2.7 & 7.9 & 3.4 & 8.0 \\
HuBERT~\citep{hsu2021hubert}            & 2.7 & 7.8 & 3.4 & 8.1 \\
data2vec~\citep{baevski2022data2vec}    & \uline{2.2} & \textbf{6.4} & \uline{2.8} & \uline{6.8} \\
\rowcolor{rowgray}
\texttt{MANAR-256.64.128}                        & 2.3 & \uline{6.7} & 2.9 & \uline{6.8} \\
\rowcolor{rowgray}
\texttt{MANAR-4K.128.128}$^\dagger$                        & \textbf{2.0} & \textbf{6.4} & \textbf{2.7} & \textbf{6.4} \\
\bottomrule
\end{tabular}

\vspace{2pt}
\caption*{\footnotesize Best results are in \textbf{bold}; second-best are \uline{underlined}.
$^{\dagger}$ Models where every third layer uses a memory size of 4K; all other layers use a memory size of 256.
}
\end{table*}

\subsection{Performance Comparison}
\label{sec:perfcomp}

\begin{figure*}[t]
\centering
\includegraphics[width=\textwidth]{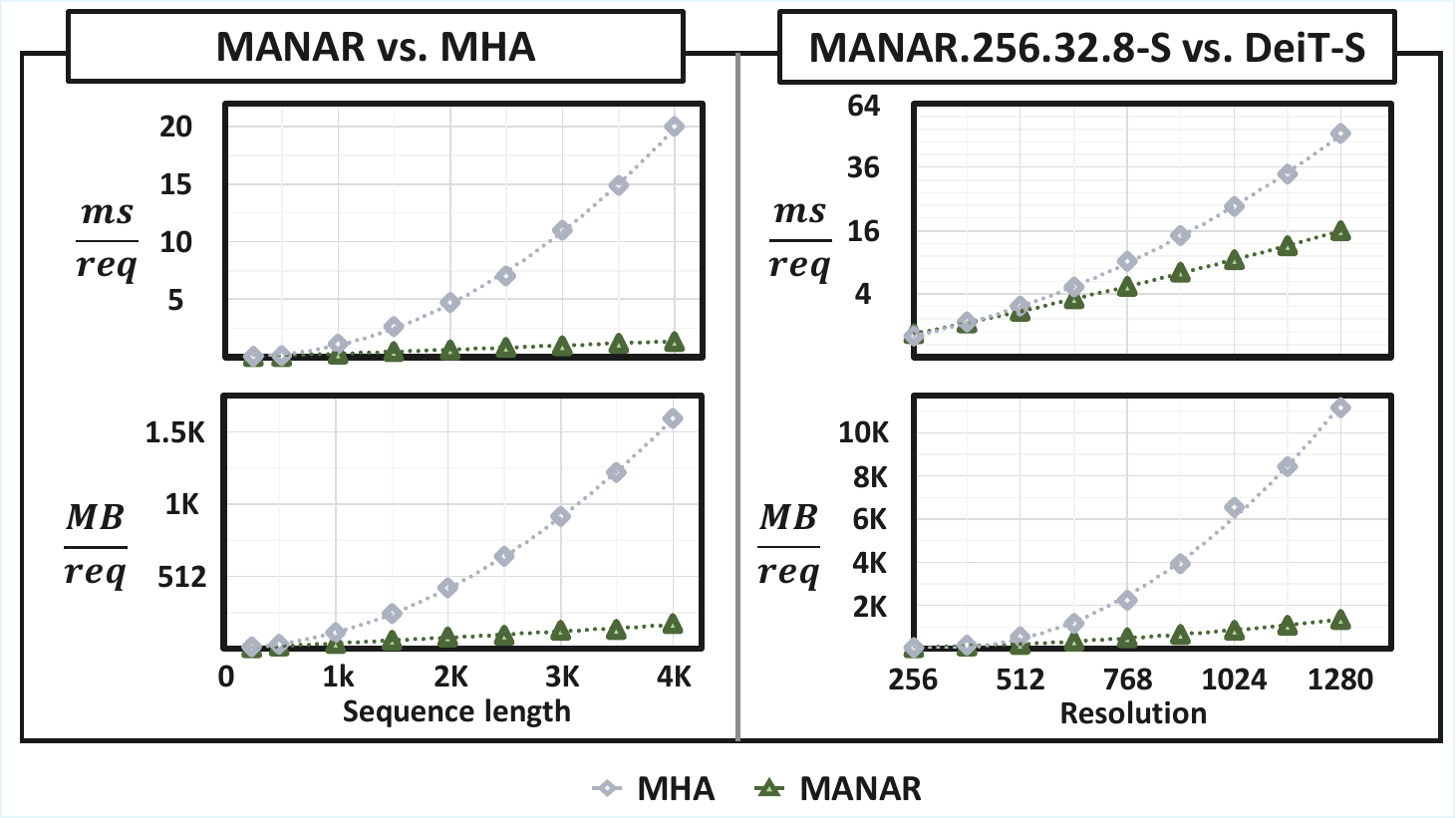}
\caption{Comparing the time and space complexity of MANAR against vanilla MHA (left), and MANAR-based DeiT-S (\texttt{MANAR-256.32.96-S}) against vanilla DeiT-S (right). The upper-right time measurements are shown on a quadratic scale to account for the resolution-dependent quadratic cost.}
\label{fig:comparison}
\end{figure*}

We profile MANAR against quadratic MHA under two conditions: single-layer throughput and an end-to-end vision model throughput. 
All measurements are performed on a single NVIDIA H100 GPU, with the batch size for each sequence length set to the maximum that could be accommodated in GPU HBM. We report wall-clock latencies averaged over repeated runs, and memory refers to peak allocated GPU memory during inference.
In the single-layer setting (left panels of Fig.~\ref{fig:comparison}), both layers use 12 heads of dimension 64, while MANAR employs 256 memory cells, an ACR of 32, and a local context window set to half the sequence length. 
At 256 tokens the two layers are essentially tied (41.5$\mu$s vs.\ 42.9$\mu$s). At 2{,}048 tokens MANAR is already $7.6\times$ faster (0.62\,ms vs.\ 4.74\,ms) while cutting peak memory $5.0\times$, and at 4{,}096 tokens the gap widens to $14.8\times$ in latency (1.35\,ms vs.\ 20.0\,ms) and $9.3\times$ in memory. 
To verify robustness, we varied the ACR size between 16 and 512 and the memory size between 256 and 16K. Speedup and memory reduction figures remained essentially unchanged across the entire range, confirming that performance and memory gains are not sensitive to these hyperparameters.

In the end-to-end setting (right panels), replacing all MHA layers in DeiT-S with \MANAR (\texttt{MANAR-256.32.96}-S) and setting local context window to half the sequence length yields slightly higher latency on $256{\times}256$ inputs (0.54\,ms vs.\ 0.44\,ms), but achieves a $2.0\times$ speed-up (7.15\,ms vs.\ 14.6\,ms) and $4.5\times$ memory reduction at $896{\times}896$, rising to $3.1\times$ faster (16.2\,ms vs.\ 49.8\,ms) and $8.2\times$ leaner at $1{,}280{\times}1{,}280$. 

These gains are an architectural consequence of \MANAR's design: each token attends only to its local context window and the constant-sized ACR, so the number of token pairs participating in contextualization is inherently smaller than in full all-to-all MHA. When the context window length is fixed to a constant, \MANAR achieves strictly linear time and memory complexity. Even under the conservative $cw\_len{=}n/2$ setting used here, the architectural reduction in attended token pairs translates into substantial wall-clock and memory advantages that grow with sequence length.


\subsection{Ablation Studies}
\label{sec:ablation}
Using the DeiT-S configuration from Sec.~\ref{sec:image_training}, we conduct ablation studies to analyze the impact of \MANAR’s key architectural components, focusing on the interplay between local contextualization, the Abstract Conceptual Representation (ACR), and the capacity of the memory used for retrieval.

We study the effect of varying the local context window length (CWL) together with the ACR size. \MANAR contextualizes each token using a combination of local neighborhood information and a global ACR constructed from retrieved memory concepts. The results show that reducing the context window degrades accuracy significantly only when the window becomes extremely small (e.g., 25\% of the sequence), indicating that purely local contextualization is insufficient in that regime. Importantly, increasing the ACR size consistently mitigates this degradation: even with reduced local context, a moderately sized ACR enables \MANAR to recover most of the lost accuracy. This suggests that the retrieved global representation provides a strong substitute for long-range token interactions, reducing the reliance on full all-to-all contextualization.

\begin{wrapfigure}{r}{0.5\columnwidth}
\vspace{-6pt}
\centering
\includegraphics[width=\linewidth]{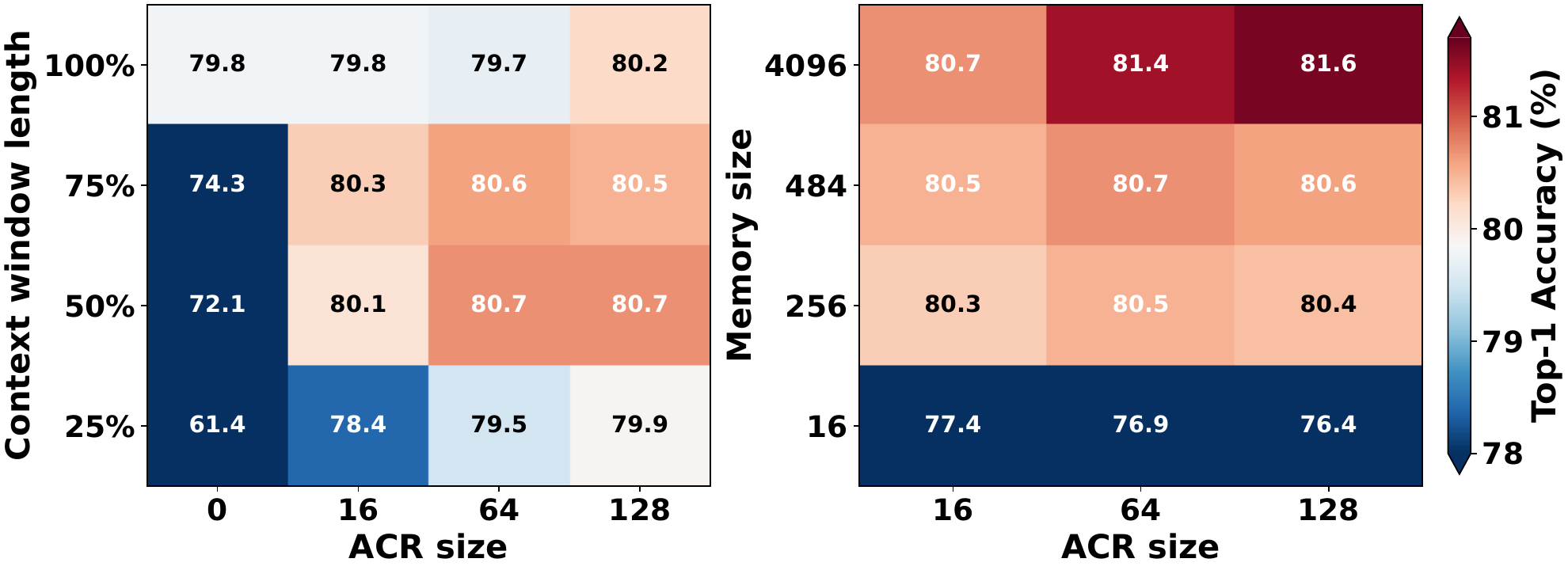}
\vspace{-8pt}
\caption{Colormap is normalized to [78, 81.7] to highlight differences in the high-accuracy regime; values outside the range saturate.}
\vspace{-6pt}
\label{fig:ablation}
\end{wrapfigure}

We next examine the impact of memory size in conjunction with ACR size. In contrast to the local context window, memory capacity exhibits a consistent trend: as memory size increases, accuracy improves steadily across all ACR sizes, with no indication of early saturation in the evaluated range. Larger memory enables \MANAR to retrieve a more diverse and informative set of concepts, which in turn leads to a more expressive and accurate ACR. Notably, while small memory severely limits performance, enlarging the memory bank continues to yield measurable gains, highlighting memory capacity as a primary driver of model accuracy.

Taken together, these ablation studies demonstrate that \MANAR benefits most from increased memory capacity, with accuracy improving consistently as more memorized concepts become available for retrieval. While local contextualization and ACR size control how retrieved information is integrated, memory size fundamentally determines the richness of the global representation that guides contextualization. This monotonic accuracy--capacity relationship supports the central design premise of \MANAR: augmenting attention with a sufficiently large, retrievable memory enables strong performance without resorting to quadratic all-to-all interactions, thereby achieving an effective trade-off between accuracy and efficiency.

\subsection{Measuring non-convex contextualization}
\label{sec:chm}

To quantify \MANAR’s ability to produce contextualized representations that are not expressible as convex combinations of attended value vectors, we adopt the Convex Hull Membership (CHM) criterion. Given a contextualized output vector $y_i \in \mathbb{R}^d$ and the set of value vectors $\{v_1,\dots,v_n\}$ involved in its contextualization, CHM tests whether $y_i$ can be expressed as a convex combination of these values. If no such combination exists, the representation lies outside the convex hull of the inputs.

We emphasize that CHM is a \emph{geometric} diagnostic of representational expansion beyond convex attention-style aggregation; we do not interpret it as a direct measure of human-like creativity or reasoning.

\begin{wrapfigure}{r}{0.45\columnwidth}
\vspace{-6pt}
\centering
\includegraphics[width=0.45\columnwidth]{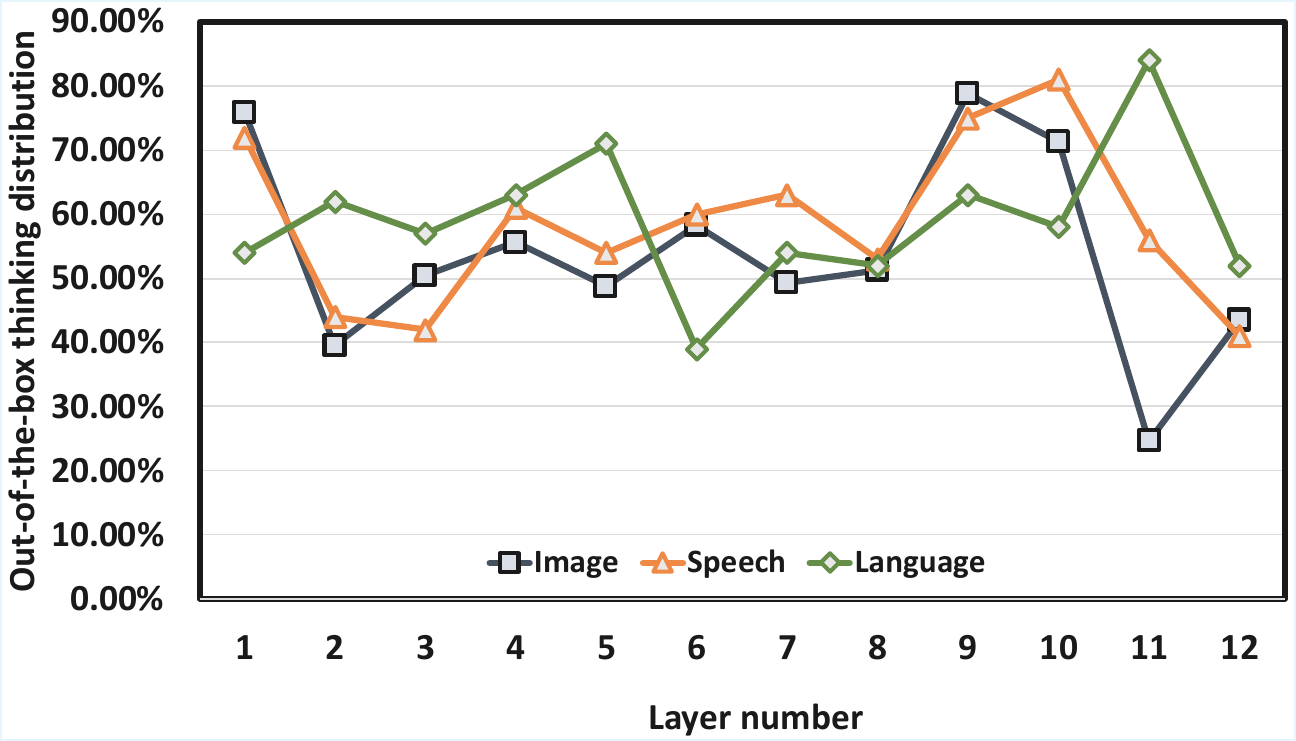}
\vspace{-8pt}
\caption{Fraction of layer outputs lying outside the convex hull of input values (CHM) across encoder layers for image, speech, and language models.}
\vspace{-6pt} 
\label{fig:ootb_exp}
\end{wrapfigure}

To ensure that CHM measurements reflect an intrinsic property of the \MANAR layer—rather than a consequence of retraining or task-specific optimization—we evaluate CHM under a strictly controlled setting in which all transformer-based weights are frozen. Specifically, \MANAR is initialized via knowledge transfer from pretrained transformer encoders, and only the additional parameters introduced by \MANAR (i.e., memory retrieval, ACR construction, and memory-to-token projection weights) are trained. For language modeling, weights are transferred from \texttt{bert-base-uncased}. For image classification and speech recognition, we use the intermediate checkpoints from the knowledge-transfer experiments in Sec.~\ref{sec:kt}, prior to unfreezing the copied weights.

This design isolates the architectural contribution of \MANAR itself. By freezing all original transformer parameters, we ensure that any deviation from convex-hull-limited representations cannot be attributed to changes in the underlying attention projections, token embeddings, or feed-forward blocks. Instead, any out-of-hull behavior must arise solely from augmenting a pretrained transformer with \MANAR’s memory. In other words, this experiment tests whether expanding MHA into \MANAR without altering the learned transformer representation space enables the model to synthesize representations that are provably unattainable by the original architecture.

CHM measurements are conducted on \texttt{MANAR-256.32.96-B} (image), \\\texttt{MANAR-256.64.128} (speech), and \texttt{MANAR-484.64.128} (language). For each encoder layer, we uniformly sample 10,000 output tokens across all heads and solve the CHM feasibility problem using linear programming. Figure~\ref{fig:ootb_exp} reports the fraction of outputs lying outside the convex hull for each layer and modality.

Across all three modalities, a large fraction of representations, often exceeding 50\%, lie outside the convex hull of the input values, confirming that \MANAR consistently produces non-convex contextualizations even when built on top of a frozen transformer. Early layers exhibit substantial out-of-hull behavior, reflecting reliance on memory-driven global context when local evidence is limited. Mid-stack layers show a temporary stabilization as token-level features consolidate. In later layers, the out-of-hull fraction rises sharply, indicating renewed use of memory concepts when synthesizing higher-level, task-specific abstractions, before declining near the output layer.

In the context of GWT, this non-convexity reflects the creative synthesis of the global workspace: the ability to generate novel abstractions by combining external stimuli with internalized prior knowledge~\cite{shanahan2006cognitive}. The significant out-of-hull behavior observed across modalities confirms that the ACR synthesizes a broader representational space than token-only aggregation, and that this behavior is a direct consequence of \MANAR’s memory-augmented architecture rather than retraining effects.

\section{Conclusion}
We introduced \textbf{MANAR}, a memory-augmented contextualization layer that generalizes standard multi-head attention by explicitly instantiating the functional principles of Global Workspace Theory (GWT). By implementing a central workspace through an Abstract Conceptual Representation (ACR) and a retrievable external memory, MANAR enables the efficient integration of global information. Crucially, we demonstrate that linear-time scaling is not merely an optimization goal but a fundamental byproduct of the GWT bottleneck, which resolves the quadratic complexity of standard attention under fixed context windows.

Across language, vision, and speech modalities, MANAR matches or exceeds strong baselines (average GLUE 85.1, 83.9\% ImageNet-1K top-1, 2.7\%/6.4\% LibriSpeech WER) while remaining a compatible re-parameterization of MHA that permits knowledge transfer from pretrained models. 

Beyond empirical performance, MANAR expands the representational capacity of the attention mechanism through non-convex contextualization. Our Convex Hull Membership (CHM) analysis confirms that the model synthesizes novel abstractions reflecting the creative synthesis of a global workspace even when grafted onto frozen pretrained transformers.

For practitioners, MANAR can be used as a drop-in replacement with knowledge transfer to reduce training cost; it is relevant to information retrieval, document and text analysis, and multimodal or speech-based systems.

Several directions remain for future work. The current \textbf{MANAR} formulation targets encoder architectures and requires adaptation to support causal, decoder-style attention. Additionally, integrating MANAR with rotary positional encodings and evaluating its scaling behavior under longer-context pre-training schedules remain important avenues for extending this brain-inspired architectural primitive. Code is available at \url{https://github.com/zuherJahshan/manar}.

\section*{Declaration of generative AI and AI-assisted technologies in the writing process}
During the preparation of this work the authors used generative AI and AI-assisted tools for rephrasing of selected text and for identifying and organizing relevant references. All content was reviewed, verified, and edited by the authors. The authors take full responsibility for the accuracy, integrity, and originality of the manuscript.








\bibliography{refs}
\end{document}